\documentclass[journal,10pt]{IEEEtran}
\usepackage{graphicx}
\usepackage{amsmath}
\usepackage{array}
\usepackage{times}
\usepackage{epsfig}
\usepackage{amssymb}
\usepackage{epstopdf}
\usepackage{float}
\usepackage{multirow}
\usepackage[footnotesize]{caption}
\usepackage{chngpage}
\usepackage{array}
\usepackage{hyperref}
\usepackage{fancyhdr}

\usepackage{xspace}

\ifCLASSINFOpdf
\else
\fi

\usepackage{url}
\begin{document}

\title{Monocular Depth Estimation with Augmented Ordinal Depth Relationships}

\author{Yuanzhouhan Cao, Tianqi Zhao, Ke Xian,  Chunhua Shen, Zhiguo Cao, Shugong Xu
\thanks{Y. Cao and C. Shen are with The University of Adelaide, SA 5005, Australia. Corresponding author: Y. Cao (e-mail:yuanzhouhan.cao@adelaide.edu.au).
}
\thanks{T. Zhao is with Tsinghua University, Beijing 100084, China.}
\thanks{K. Xian and Z. Cao are with Huazhong University of Science and Technology, Wuhan 430074, China. K. Xian's contribition was made when he was visiting The University of Adelaide.}
\thanks{S. Xu  is with the Shanghai Institute for Advanced Communications and Data Science, Shanghai University, Shanghai 200444, China.}
}

\maketitle

\thispagestyle{fancy}
\fancyhead{}
\chead{}
\rhead{}
\lfoot{}
\cfoot{\thepage}   %
\rfoot{}
\fancyhead[L]{Preprint}
\renewcommand{\headrulewidth}{1pt}

\begin{abstract}
Most existing algorithms for depth estimation from single monocular images need large quantities of metric ground-truth depths for supervised learning. We show that relative depth can be an informative cue for metric depth estimation and can be easily obtained from vast stereo videos. Acquiring metric depths from stereo videos is sometimes impracticable due to the absence of camera parameters. In this paper, we propose to improve the performance of metric depth estimation with relative depths collected from stereo movie videos using existing stereo matching algorithm. We introduce a new ``Relative Depth in Stereo" (RDIS) dataset densely labelled with relative depths. We first pretrain a ResNet model on our RDIS dataset. Then we finetune the model on RGB-D datasets with metric ground-truth depths. During our finetuning, we formulate depth estimation as a classification task. This re-formulation scheme enables us to obtain the confidence of a depth prediction in the form of probability distribution. With this confidence, we propose an information gain loss to make use of the predictions that are close to ground-truth. We evaluate our approach on both indoor and outdoor benchmark RGB-D datasets and achieve state-of-the-art performance.
\end{abstract}


\IEEEpeerreviewmaketitle

\section{Introduction}
Predicting accurate depths from single monocular images is a fundamental task in computer vision and has been an active research topic for decades. Typical methods formulate depth estimation as a supervised learning task~\cite{Saxena2005,liu2015CVPR,EigenPF14}. As a result, large amounts of metric ground-truth depths are needed. However, the acquisition of metric ground-truth depths requires depth sensors, and the collected RGB-D training data is limited in the size as well as the diversity of scenes due to the limitation of depth sensors. For example, the popular Microsoft Kinect can not obtain the depths of far objects in outdoor scenes.

In order to overcome the problem of limited metric ground-truth depths, some recent works manage to predict depths from stereo videos~\cite{garg2016unsupervised,godard2016unsupervised,xie2016deep3d} without the supervision of ground-truth depths. Specifically, the model is trained by computing the disparity maps and minimizing an image reconstruction loss between training stereo pairs. The performance is not satisfactory due to the absence of ground-truth depths during training. However, the training stereo videos are easier to obtain than metric ground-truth depths and are plenty in terms of amount as well as scene diversity.

Driven by the aforementioned characteristics of recent depth estimation methods, a question arises: Is it possible to acquire large quantities of training data from stereo videos to improve the performance of monocular depth estimation?

\begin{figure}
	\begin{center}
		\includegraphics[scale=.50]{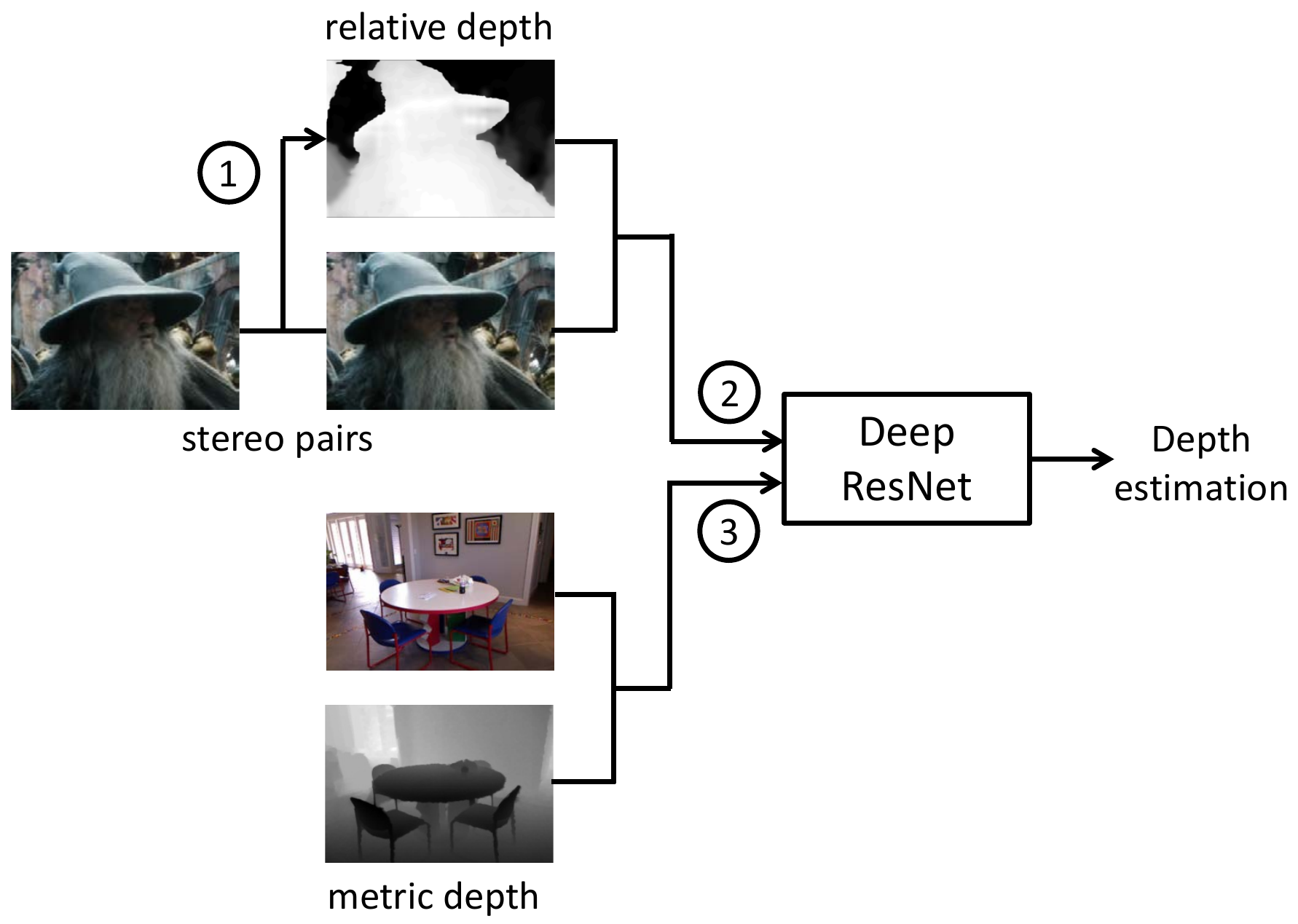}
	\end{center}
	\caption{Overview of our proposed depth estimation method. We first generate relative depths from stereo pairs, then pretrain a deep residual network with the relative depths. Finally, we finetune the network with metric depths for monocular depth estimation.}
	\label{fig_overview}
\end{figure}

Compared to metric depths, relative depths can be easily obtained from stereo videos using existing stereo matching algorithms~\cite{mccnn2016stereo,Spyropoulos14,Luo_2016_CVPR,ZhangLL09}. Models trained with abundant relative depths are able to estimate dense relative depth maps. Although the values are not metric depth values, they provide two types of important information: (a) global layout of the scene. (b) relative relationships of points. These information work as priors in inferring metric depths. The recent works by Zoran et al.~\cite{zoran2015learning} and Chen et al.~\cite{DIW_NIPS2016} have revealed that it is possible to predict satisfactory metric depths with only relative ground-truth depths. In this paper, we propose to improve the performance of metric depth estimation with relative depths generated from stereo movie videos. An overview of our approach is illustrated in Fig.~\ref{fig_overview}. Our approach can be broadly divided into 3 steps: We first obtain ground-truth relative depths from stereo movie videos, then we pretrain a deep residual network with our relative ground-truth depths. Finally, we finetune our network on benchmark RGB-D datasets with metric ground-truth depths. Note that, as we exploit 3D movie stereo videos, which do not have the camera parameters and typically are re-calibrated for display, it is impossible to compute the metric depth.

Most existing methods formulate depth estimation as a regression problem due to the continuous property of depths~\cite{liu2015CVPR,EigenPF14,laina2016deeper}. For human beings, we may find it difficult to tell the exact distance of a specific point in a natural scene, but we can easily give a rough distance range of that point.
Motivated by this, we formulate depth estimation as a pixel-wise classification task by discretizing the continuous depth values into several discrete bins and show that this simple re-formulation scheme performs surprisingly well. More importantly, we can easily obtain the confidence of a depth prediction in the form of probability distribution. With this confidence, we can apply an information gain loss to make use of the predictions that are close to ground-truth during training.

To summarize, we highlight the contributions of our work as follows:
\begin{enumerate}
\item
We formulate depth estimation as a classification task and propose an information gain loss.
\item
We propose a new dataset Relative Depth in Stereo (RDIS) containing images labelled with dense relative depths. The relative depths are generated with very low cost.
\item
We show that our proposed RDIS dataset can improve the performance of metric depth estimation significantly and we outperform state-of-the-art depth estimation methods on both indoor and outdoor benchmark RGB-D datasets.

\end{enumerate}

\section{Related Work}
Traditional depth estimation methods are mainly based on geometric models. For example, the works of
\cite{Hedau,NIPS2010_4120,SchwingECCV2012} rely on box-shaped models and try to fit the box edges to those observed in the image.
These methods are limited to only model particular scene structures and therefore are not applicable for general-scene depth estimations. For example, Saxena et al.~\cite{Saxena2009} devised a multi-scale MRF, but assumed that all scenes were horizontally aligned with the ground plane. Hoiem et al.~\cite{Hoiem_2005},  instead of predicting depth explicitly, estimated geometric structures for major image regions and
composed simple 3D models to represent the scene. More recently, non-parametric methods \cite{KKarsch} are explored. These methods consist of candidate images retrieval, scene alignment and then depth inference using optimizations with smoothness constraints. These methods are based on the assumption that scenes with semantically similar appearances should have similar depth distributions when densely aligned.

Most depth estimation algorithms in recent years achieve outstanding performance by training deep convolutional neural networks (CNN)~\cite{NIPS2012_4824,Simonyan14c,long_shelhamer_fcn} with fully annotated RGB-D datasets~\cite{Silberman_ECCV12,Geiger2013IJRR}. Liu et al.~\cite{LiuSLR15} presented a deep convolutional neural field which jointly learns the unary and pairwise potentials of continuous conditional random fields (CRF) in a unified deep network. Eigen et al.~\cite{Eigen15} proposed a multi-scale network architecture to predict depths as well as surface normals and semantic labels. Li et al.~\cite{LiB15} and Wang et al.~\cite{Wang_2015_CVPR} formulated depth estimation in a two-layer hierarchical CRF to enforce synergy between global and local predictions. Laina et al.~\cite{laina2016deeper} applied the latest deep residual network~\cite{kmhe15} as well as an up-sampling scheme for depth estimation. Li et al.~\cite{junli_depth} proposed a two-streamed CNN that predicts depth and depth gradients, which are fused into an accurate depth map. Chakrabarti et al.~\cite{NIPS2016_6510} proposed a CNN which is able to capture cues informative towards different aspects of local geometric structure. Xu et al.~\cite{danxu_2017} proposed a deep model which fuses complementary information derived from multiple CNN side outputs.

Other recent works managed to train deep CNNs for depth estimation in an unsupervised manner. To name a few, Garg et al.~\cite{garg2016unsupervised} and Cl{\'e}ment et al.~\cite{godard2016unsupervised} treated depth estimation as an image reconstruction~\cite{flynn2015deepstereo} problem during training and output disparity maps during prediction. In order to construct a fully differentiable training loss, Taylor approximation and bilinear interpolation are applied in~\cite{garg2016unsupervised} and~\cite{godard2016unsupervised} respectively. Since the network outputs of~\cite{garg2016unsupervised} and~\cite{godard2016unsupervised} are disparity maps, camera parameters are needed to recover the metric depths. Similarly, the Deep3D model~\cite{xie2016deep3d} also applied an image reconstruction loss during training, where their goal is to predict the right view from the left view of a stereo pair, and the disparity map is generated internally.

Ordinal relationships and rankings have also been exploited for mid-level vision tasks including depth estimation in recent years. Zoran et al.~\cite{zoran2015learning} learned the ordinal relationships between pairs of points using a classification loss, then they solved a constrained quadratic optimization problem to map the ordinal estimates to metric values. Chen et al.~\cite{DIW_NIPS2016} proposed to learn ordinal relationships through a ranking loss~\cite{cao2007learning} and retrieve the metric depth values by simple normalization. Notably,~\cite{DIW_NIPS2016} also proposed a new dataset Depth in the Wild (DIW) consisting of images in the wild labelled with relative depths.

Our work is mainly inspired by Chen et al.'s single-image depth perception in the wild~\cite{DIW_NIPS2016}. However, our approach is different in three distinct aspects. First, instead of manually labelling pixels with relative relationships, we acquire relative depths using existing stereo matching algorithm from stereo movie videos and thus can obtain large amount of training data with low cost. Second, instead of labelling only one pair of points per image with relative relationships, we generate dense relative depth maps. Finally, in order to retrieve metric depth predictions, they arbitrarily normalize the predicted relative depth maps such that the mean and standard deviation are the same with the metric ground-truth depths of training set, while we finetune our pretrained network with metric ground-truth depths for better performance.

\section{Proposed Method}
In this section, we elaborate our proposed method for monocular depth estimation. We first present the stereo matching algorithm that we used to generate relative ground-truth depth. Then we introduce our network architecture, followed by the introduction of our loss functions.

\subsection{Relative depth generation}

\begin{figure*}
	\begin{center}
		\includegraphics[scale=.51]{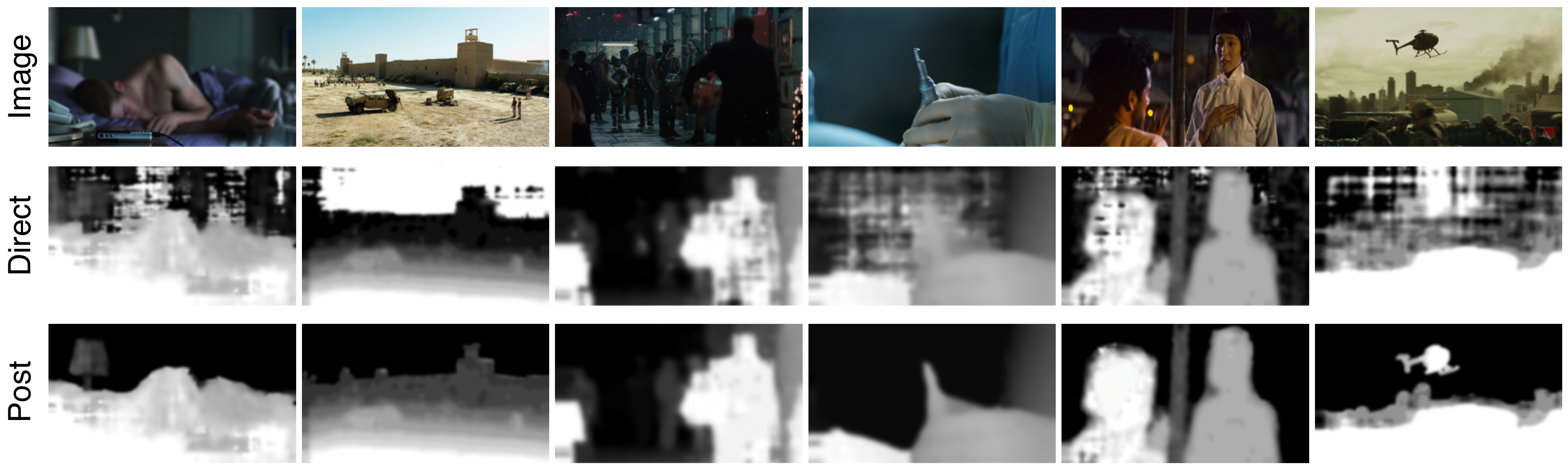}
	\end{center}
	\caption{Some examples of our Relative Depth in Stereo (RDIS) dataset. The first row are RGB images, the second row are disparity maps directly generated by stereo algorithm. The last row are post-processed disparity maps, which are used as ground-truths.}
	\label{fig_BJ3Dsample}
\end{figure*}

The first step of our approach is to generate relative ground-truth depth from stereo videos using existing stereo matching algorithm. Stereo matching algorithms rely on computing matching costs to measure the similarities of stereo pairs. In this paper, we choose the commonly-used absolute difference (AD) matching cost combined with a background subtraction by bilateral filtering (BilSub) which has been proven to perform well by Hirschmuller et al.~\cite{Heikostereomatch}. For a pixel $\mathnormal{p}$ in the left image, its corresponding pixel in the right image is represented as $\mathnormal{p-d}$, where $\mathnormal{d}$ is the disparity. The absolute difference is represented as:
\begin{equation} \label{ad}
\mathnormal{C_{ad}(p,d)}=|I_L(p)-I_R(p-d)|,
\end{equation}
where $\mathnormal{I_L}$ and $\mathnormal{I_R}$ are left and right images respectively. We sum the costs of all three channels of color images. The bilateral filtering effectively removes a local offset without blurring high contrast texture differences that may correspond to depth discontinuities.

As for the stereo algorithm, we use the semi-global matching (SGM) method~\cite{Hirschmuller_2008_MI}. It aims to minimize a global 2D energy function by solving a large number of 1D minimization problems. The energy function is:
\begin{small}
\begin{equation}\label{stereo_energ}
\begin{aligned}
\mathnormal{E(D)}=&\sum_{p}(C(p,D_p)+\sum_{q \in N_p}P_1 \mathrm{T}[|D_p-D_q|=1] \\
&+\sum_{q \in N_p}P_2 \mathrm{T}[|D_p-D_q|>1]),
\end{aligned}
\end{equation}
\end{small}where the first term calculates the sum of a pixel-wise matching cost for all pixels at their disparities $D_p$. The second term adds a constant penalty $P1$ for all pixels $q$ in the neighborhood $N_p$ of $p$, for which the disparity changes a little bit (i.e., 1 pixel). Similarly, the third term adds a larger constant penalty $P2$, for all larger disparity changes. The SGM calculates $E(D)$ along 1D paths from 8 directions towards each pixel of interest using dynamic programming. The costs of all paths are summed for each pixel and disparity. The disparity is then determined by winner-takes-all. During training, We label the pairs of points with ordinal relationships (farther, closer, equal) according to their disparities. Since the disparity values of two points can not be exactly the same, we apply a relaxed definition of equality. The ordinal relationship of a pair of points is equal if the disparity difference is smaller than a fixed threshold.

The direct output of stereo matching algorithm is a dense disparity map with the left image treated as the reference image. This disparity map can not be directly used for training due to the defects such as noise, discontinuities or incorrect values. Some examples are shown in Fig.~\ref{fig_BJ3Dsample}. As a result, we need to post-process the disparity maps generated by stereo algorithm. The post-processing is done by experienced workers from movie production company using professional movie production software. Specifically, we first correct the vague or missing boundaries of objects using B-splines, then we smooth the disparity values within objects and background. It takes a median of 90 seconds to post-process an image of our dataset. Although the labelling of our dataset takes longer time than the DIW dataset, our dataset is densely labelled and contains more ordinal relationships than the DIW dataset. In terms of single ordinal relationship labelling our method is much more efficient. After post-processing, each disparity map is visually checked by two workers according to the intensities. The workers are required to assign ``overall correct", ``contain mislabelled parts" or ``not sure" to each disparity map. We only keep the disparity maps which both workers assigned as ``overall correct".

We also test several other stereo matching algorithms including the deep learning based. Although the qualities of these direct output disparities are different, they are all very coarse, furthermore the difference becomes negligible after human post-processing. So we pick the simplest stereo matching method.

We collect 70 3D movies produced in recent years. Since the stereo matching algorithm requires the stereo videos to be rectified, we only use 3D movies created by post-production instead of movies taken with stereo cameras. In order to avoid similar frames, we only select roughly 1500 frames in each movie. With the selected frames, we generate a new dataset Relative Depth in Stereo (RDIS) containing 97652 training images labelled with dense relative ground-truth depths. Notably, we can not obtain the metric depths from the relative depths because we do not have the camera parameters of these 3D movies. Our dataset has no test images because: 1) The goal of our dataset is to improve the performance of metric depth estimation. 2) Our ground-truth disparities are obtained through stereo matching algorithm and inevitably contain noisy points.

\subsection{Network architecture}
Recently, a deep residual learning framework has been introduced by He et al.~\cite{kmhe15,kmhe16eccv} and showing compelling accuracy and nice convergence behaviours. In our work, we follow the deep residual network architecture proposed by Wu et al.~\cite{zifengwide} which contains fewer layers but outperforms the deep residual network with 152 layers in \cite{kmhe15}.

Instead of directly learning the underlying mapping of a few stacked layer, the deep residual network learns the residual mapping through building blocks. We consider two types of building blocks in our network architecture. The first is defined as:

\begin{equation} \label{block1}
\mathbf{y} = F(\mathbf{x},\{W_i\})+\mathbf{x},
\end{equation}
where $\mathbf{x}$ and $\mathbf{y}$ are the input and output matrices of stacked layers respectively. The function $F(\mathbf{x},\{W_i\})$ is the residual mapping that need to be learned. The dimensions of $\mathbf{x}$ and $F$ need to be equal since the addition is element-wise. If this is not the case, we apply another building block defined as:

\begin{equation}
\mathbf{y} = F(\mathbf{x},\{W_i\})+W_s\mathbf{x}.
\end{equation}
Comparing to the shortcut connection in Eq.~(\ref{block1}), a linear projection $\mathnormal{W_s}$ is applied to match the dimensions of $\mathbf{x}$ and $F$.

\begin{figure}
	\begin{center}
		\includegraphics[scale=.351]{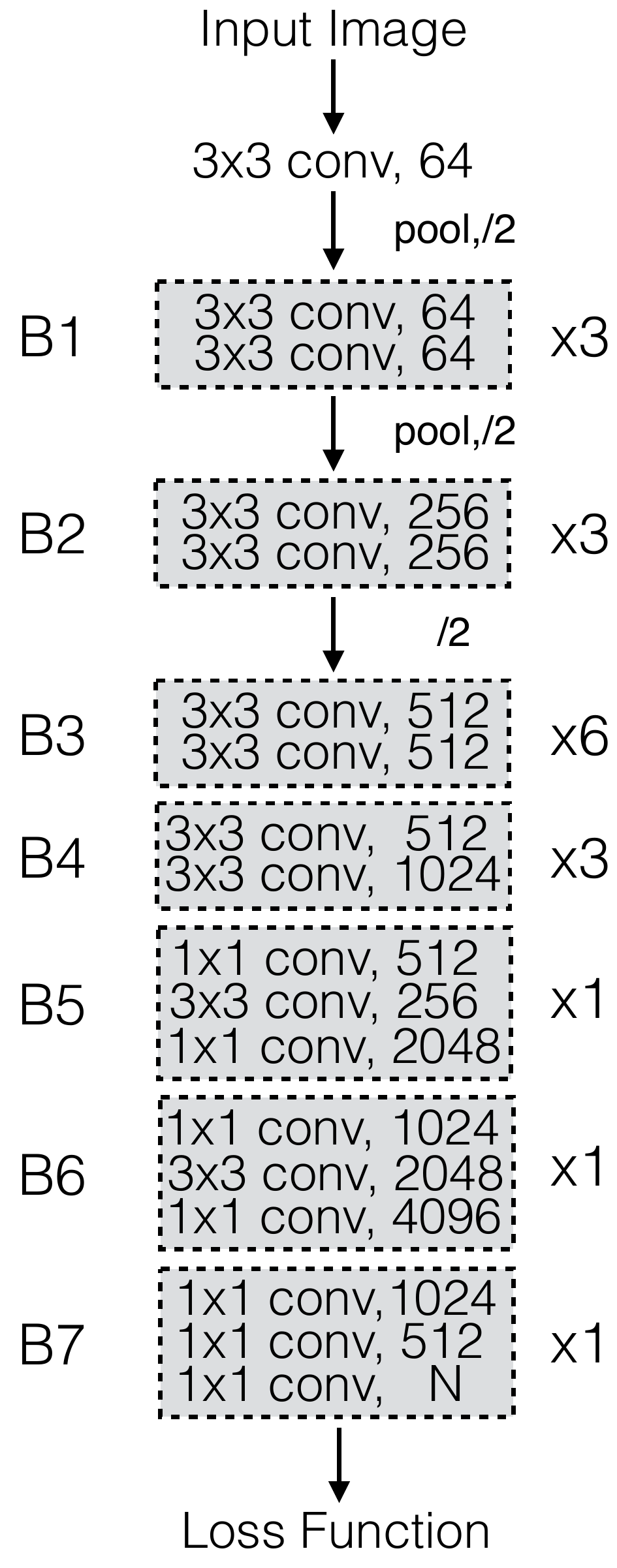}
	\end{center}
	\caption{Detailed structure of our deep residual network. It has 7 convolution blocks, each with different numbers of residual units.}
	\label{fig_network}
\end{figure}

We illustrate our detailed network structure in Fig. \ref{fig_network}. Generally, it is composed of 7 convolution blocks. Two max pooling layers with stride of 2 are applied before the first and the second convolution blocks. The first convolutional layers of block 3, 4 and 5 have a stride of 2. Batch normalizations (BNs)~\cite{batchnorm} and ReLUs are applied before weight layers. We initialize the layers up to block 6 with our model pretrained on the ImageNet~\cite{ILSVRC15} and Places365~\cite{places365} datasets. After block 6, we add 3 convolutional layers with randomly initialized weights. The channels of the first and second added convolutional layers are 1024 and 512 respectively. The channel number of last convolutional layer is determined by the loss function. The channel number is 1 for the pretraining using ranking loss. As for the finetuning, we discretize the continuous metric depths into several bins and formulate depth estimation as a classification task, the channel number is equal to the bin number. We give more details about the loss functions below.

\subsection{Loss function}
Our proposed approach for depth estimation contains two training stages: pretraining with relative depths and finetuning with metric depths. For the pretraining, we employ the ranking loss which encourages a small difference between depths if the ground-truth ordinal relation is equality and encourages a large difference otherwise. Specifically, consider a training image $\mathnormal{I}$ with $\mathnormal{K}$ pairs of points with ground-truth ordinal relations $\mathnormal{R}=\{(i_k,j_k,r_k)\},k\in[1,\dots,\mathnormal{K}],$ where $i_k$ and $j_k$ are the two points of $k$-th pair, and $r_k$ is the ground-truth depth relation between $i_k$ and $j_k$: closer ($+1$), farther ($-1$) and equal ($0$). Let $z$ be the output depth map of our deep residual network and $z_{i_k},z_{j_k}$ be the predicted depth values of $i_k$ and $j_k$. The ranking loss is defined as:
\begin{equation} \label{rankloss1}
\mathnormal{L_{rank}(I,R,z)}=\sum_{k=1}^{K}E(I,i_k,j_k,r,z),
\end{equation}
where $E(I,i_k,j_k,r,z)$ is the loss of the $k$-th pair:
\begin{equation} \label{rankloss2}
E =
\begin{cases}
\log(1+\exp(-z_{i_k}+z_{j_k})),&r_k=+1;\\
\log(1+\exp(z_{i_k}-z_{j_k})),&r_k=-1;\\
(z_{j_k}-z_{i_k})^2,      &r_k=0.
\end{cases}
\end{equation}

After pretraining, we finetune our network with discretized metric depths. We use the pixel-wise multinomial logistic loss defined as:
\begin{equation} \label{logloss}
\mathnormal{L_{\log}}=-\frac{1}{N}\sum_{i=1}^{N}\sum_{D=1}^{B}H(D_{i}^{*},D)\log(\mathnormal{P(D|z_i)}),
\end{equation}
where $\mathnormal{D_{i}^{*}}\in[1,\dots,B]$ is the ground-truth depth label of pixel $\mathnormal{i}$ and $B$ is the total number of discretization bins. $\mathnormal{N}$ is the number of pixels. $\mathnormal{P(D|z_i)} = {e^{z_i,D}}/{\sum_{d=1}^{B}{e^{z_{i,d}}}}$ is the probability of pixel $\mathnormal{i}$ labelled with $\mathnormal{D}$. $\mathnormal{z_{i,d}}$ is the output of last convolutional layer in the network.

Although we formulate depth estimation as a classification task by discretizing continuous depth values into several bins, the depth labels are different with the labels of other classification tasks (e.g., semantic segmentation). Predicted depth labels that are closer to ground-truth should have more contribution in updating network weights. This is achieved through the information gain matrix $\mathnormal{H}$ in Eq.~(\ref{logloss}). It is a $B \times B$ symmetric matrix with elements $H(p,q) = \exp[-\alpha(p-q)^2]$ and $\alpha$ is a constant. During training, we equally discretize the continuous depths in the log space into several bins and during prediction, we set the depth value of each pixel to be the center of its corresponding bin.

\section{Experiments}
We organize our experiments into the following 3 parts: 1) We demonstrate the benefit of the pretraining on our proposed Relative Depth in Stereo (RDIS) dataset by comparing with other pretraining schemes; 2) We evaluate the metric depth estimation on indoor and outdoor benchmark RGB-D datasets and analyze the contributions of some key components in our approach; 3) We evaluate both metric and relative depth estimation and compare with state-of-the-art results. During pretraining and finetuning, we apply online data augmentation including random scaling and flipping. We apply the following measures for metric depth evaluation:

$\bullet$ root mean squared error (rms): $\sqrt{\frac{1}{T}\sum_{p}(d_{gt}-d_p)^2}$

$\bullet$ average relative error (rel): $\frac{1}{T}\sum_{p}\frac{|d_{gt}-d_{p}|}{d_{gt}}$

$\bullet$ average $\log_{10}$ error (log10): $\frac{1}{T}\sum_{p}|\log_{10}d_{gt} - \log_{10}d_p|$

$\bullet$ root mean squared log error (rmslog) $\sqrt{\frac{1}{T}\sum_{p}(\log d_{gt} - \log d_p)^2}$

$\bullet$ accuracy with threshold $thr$:

percentage ($\%$) of $d_p$ s.t. $\max(\frac{d_{gt}}{d_p},\frac{d_p}{d_{gt}}) = \delta < thr$ where $d_{gt}$ and $d_p$ are the ground-truth and predicted depths respectively of pixels, and $T$ is the total number of pixels in all the evaluated images. As for the relative depth evaluation, we report the Weighted Human Disagreement Rate (WHDR)~\cite{zoran2015learning},  the average disagreement rate with human annotators, weighted by their confidence (here set to 1).

We implement our network training based on the MXNet~\cite{ChenLLLWWXXZZ15}. During pretraining, we set the batch size to be 16 and the initial learning rate to be 0.0006. We train for 40k iterations and decrease the learning rate by a factor of 10 at 25k and 35k iterations. During finetuning, all layers are optimized. We set the batch size to be 4 and the initial learning rate to be 0.0004. We ran 15k iterations on the standard NYUD2 dataset and decrease the learning rate by a factor of 10 at 10k iterations. We ran 160k iterations on the raw NYUD2 dataset and decrease the learning rate by a factor of 10 at 90k and 120k iterations. We ran 9k iterations on the standard KITTI dataset and decrease the learning rate by a factor of 10 at 6k iterations. We ran 45k iterations on the raw KITTI dataset and decrease the learning rate by a factor of 10 at 30k and 40k iterations.

\subsection{Benefit of pretraining}\label{exp_pretrain}
In this section, we show the benefit of the pretraining with our proposed RDIS dataset. Since our proposed RDIS dataset is densely labelled with relative depths, we need first to determine the number of ground-truth pairs in each image during pretraining. We randomly sample 100, 500, 1K and 5K ground-truth pairs in each input image during pretraining and finetune on both the NYUD2~\cite{Silberman_ECCV12} and KITTI~\cite{Geiger2013IJRR} datasets.

The standard NYUD2 training set contains 795 images. We split the 795 images into a training set with 400 images and a validation set with 395 images. We discretize the continuous metric depth values into 100 bins in the log space. As for the KITTI dataset, we apply the same split in~\cite{EigenPF14} which contains 700 training images and 697 test images. We further evenly split the 700 training images into a training set and a validation set. We only use left images and discretize the continuous metric depth values into 50 bins in the log space. We cap the maximum depth to be 80 meters. For both the NYUD2 and KITTI datasets, we finetune on our split training sets and evaluate on our validation sets. During finetuning, we ignore the missing values in ground-truth depth maps and only evaluate on valid points. We do not apply the information gain matrix in this experiment. The results are illustrated in Table~\ref{table:number_of_pairs}. As we can see from the table that for both indoor and outdoor datasets, the performance increases with the number of pairs and achieves the best when using 1K pairs per input image. Further increasing the number of pairs does not improve the performance. For pretraining with 5K pairs of points, we further add dropouts and evaluate the accuracy with $\delta<1.25$ on the NYUD2 dataset. The accuracies are 63.7\%, 63.1\%, 62.6\% and 62.3\% with 32K,
35K, 40K and 43K iterations. It demonstrates that the performance decrease is caused by overfitting. In the following experiments, we all sample 1K ground-truth pairs in each input image during pretraining.

\begin{table}
	\renewcommand\arraystretch{1.2}
	\caption{Comparison between different numbers of pairs during pretraining. The model is pretrained on our RDIS dataset and finetuned on NYUD2 and KITTI datasets. For each dataset, each row represents different numbers of ground-truth pairs in each input image during pretraining. }
	\centering
	\label{table:number_of_pairs}
	\begin{tabular}{@{\hskip 0.12cm}c@{\hskip 0.12cm}c@{\hskip 0.12cm}c@{\hskip 0.12cm}c @{\hskip 0.42cm}c@{\hskip 0.15cm}c@{\hskip 0.15cm}c}
        \noalign{\smallskip}
		\hline
		\noalign{\smallskip}
						& \multicolumn{3}{c}{\small Accuracy}    & \multicolumn{3}{c}{\small Error}  \\
		                 &\small $\delta<1.25$   &\small $\delta<1.25^2$    &\small $\delta<1.25^3$    &\small rel   &\small log10  &\small rms  \\
		\noalign{\smallskip}
		\hline
		\noalign{\smallskip}
		& \multicolumn{5}{c}{\small NYUD2}     \\
		\hline
		\noalign{\smallskip}
   \small 100 pairs         &\small 63.8\%   &\small 90.2\%   &\small 97.5\%   &\small 0.202    &\small 0.090    &\small 0.816  \\
		\noalign{\smallskip}
   \small 500 pairs         &\small 68.2\%   &\small  92.5\%  &\small  98.6\%  &\small  0.178   &\small  0.081   &\small  0.750  \\
		\noalign{\smallskip}
   \small 1K pairs         &\small \bf 71.1\%   &\small \bf 93.3\%   &\small \bf 98.6\%   &\small \bf 0.173    &\small \bf 0.077    &\small \bf 0.721  \\
		\noalign{\smallskip}
   \small 5K pairs         &\small 63.0\%   &\small 89.1\%   &\small 97.1\%   &\small 0.208    &\small 0.092    &\small 0.828  \\
		\hline
		\noalign{\smallskip}
		& \multicolumn{5}{c}{\small KITTI}     \\
		\hline
		\noalign{\smallskip}
   \small 100 pairs   &\small  70.6\%   &\small 88.3\%   &\small 94.6\%   &\small 0.230    &\small  0.088   &\small 6.357  \\
		\noalign{\smallskip}
   \small 500 pairs   &\small  74.1\%   &\small 90.0\%   &\small 95.3\%   &\small 0.205    &\small 0.079    &\small 5.900 \\
		\noalign{\smallskip}
   \small 1K pairs    &\small \bf 74.2\%   &\small \bf 90.0\%   &\small \bf 95.5\%   &\small \bf 0.205    &\small \bf 0.079    &\small \bf 5.828  \\
		\noalign{\smallskip}
   \small 5K pairs   &\small  70.2\%   &\small 87.3\%   &\small 94.1\%   &\small 0.223    &\small 0.088    &\small 6.629  \\
		\hline
		\noalign{\smallskip}
	\end{tabular}
\end{table}

In order to demonstrate the quality of our proposed RDIS dataset, we conduct experimental comparisons against several pretraining shcemes: 1) Directly finetune our ResNet model on RGB-D datasets without pretraining (Direct); 2) Pretrain our ResNet model on the DIW~\cite{DIW_NIPS2016} dataset and finetune on RGB-D datasets (DIW); 3) Pretrain our ResNet model using our RDIS images and finetune on RGB-D datasets, the ground-truth relative depths for pretraining are generated using the Deep3D model~\cite{xie2016deep3d} (Deep3D).

We finetune on the standard training set of the NYUD2 which contains 795 images and evaluate on the standard test set which contains 654 images. The continuous metric depth values are discretized into 100 bins in the log space. The parameter $\alpha$ of the information gain matrix defined in Eq.~(\ref{logloss}) is set to $2.0$. As for the KITTI dataset, we finetune on the same 700 training images and evaluate on the same 697 test images as in~\cite{EigenPF14}. The continuous metric depth values are discretized into 50 bins in the log space and the maximum depth value is capped to be 80 meters. The parameter $\alpha$ is set to $0.2$. We ignore the missing ground-truth values during both finetuning and evaluation.

We show the results in Table~\ref{table:pretrain_benifits}. We can see from the table that the pretraining on our proposed RDIS dataset improves the depth estimation of both indoor and outdoor datasets significantly, and even outperforms the pretraining on the DIW dataset. Notably, compared to the DIW dataset which contains 421K training images with manually labelled relative depths, our RDIS dataset contains only 97652 images, and the relative ground-truth depths are generated by existing stereo matching algorithm.

Note that in this experiment, the relative depths generated by the Deep3D model are not post-processed. In order to make a fair comparison, we post-process 5000 depth maps and conduct the experiment on the NYUD2 dataset again. With the Deep3D stero matching, we achieve 74.1\% of the 1.25 accuracy and 1.76 of the relative error. With our selected stereo matching, we achieve 74.5\% of the 1.25 accuracy and 1.79 of the relative error. This experiment shows that after human post-processing, different stereo matching algorithms have negligible difference.

\begin{table}
	\renewcommand\arraystretch{1.2}
	\caption{Test results on the NYUD2 and KITTI datasets with different pretraining. For each dataset, the first row is the result without pretraining; the second row is the result with pretaining on the DIW dataset; the third row is the result with pretaining using our RDIS images but the ground-truth relative depths are generated by the Deep3D~\cite{xie2016deep3d} model; the last row is the result with pretraining on our RDIS dataset.}
	\centering
	\label{table:pretrain_benifits}
	\begin{tabular}{@{\hskip 0.05cm}c@{\hskip 0.10cm}c@{\hskip 0.10cm}c@{\hskip 0.10cm}c @{\hskip 0.31cm}c@{\hskip 0.15cm}c@{\hskip 0.15cm}c}
        \noalign{\smallskip}
		\hline
		\noalign{\smallskip}
						& \multicolumn{3}{c}{\small Accuracy}    & \multicolumn{3}{c}{\small Error}  \\
		                 &\small $\delta<1.25$   &\small $\delta<1.25^2$    &\small $\delta<1.25^3$    &\small rel    &\small log10  &\small rms  \\
		\noalign{\smallskip}
		\hline
		\noalign{\smallskip}
		& \multicolumn{5}{c}{\small NYUD2}     \\
		\hline
		\noalign{\smallskip}
   \small Direct    &\small 73.3\%      &\small 93.5\%   &\small 98.1\%   &\small 0.186   &\small 0.075   &\small 0.666  \\
		\noalign{\smallskip}
      \small DIW        &\small 77.3\%      &\small  95.4\%   &\small  98.9\%   &\small 0.160   &\small 0.066   &\small \bf 0.600 \\
		\noalign{\smallskip}
    \small  Deep3D    &\small 72.5\%      &\small 92.8\%   &\small 97.8\%   &\small 0.191   &\small 0.077   &\small 0.683  \\
		\noalign{\smallskip}
      \small  Ours      &\small \bf 78.1\%   &\small \bf 95.4\%   &\small \bf 98.9\%   &\small \bf 0.157    &\small \bf 0.065    &\small 0.604  \\
		\hline
		\noalign{\smallskip}
		& \multicolumn{5}{c}{\small KITTI}     \\
		\hline
		\noalign{\smallskip}
		\small Direct    &\small 77.3\%      &\small 92.1\%   &\small 96.9\%   &\small 0.173   &\small 0.070   &\small 5.890  \\
		\noalign{\smallskip}
        \small DIW      &\small 79.7\%      &\small 93.7\%   &\small  97.8\%   &\small  0.154   &\small 0.064   &\small  5.251 \\
		\noalign{\smallskip}
        \small Deep3D   &\small 76.1\%      &\small 91.9\%   &\small 97.1\%   &\small 0.178   &\small 0.072   &\small 5.765  \\
		\noalign{\smallskip}
       \small  Ours      &\small \bf 82.9\%   &\small \bf 94.3\%   &\small \bf 98.2\%   &\small \bf 0.142    &\small \bf 0.058    &\small \bf 5.066  \\
		\hline
	\end{tabular}
\end{table}

\subsection{Component analysis}
In this section, we evaluate metric depth estimation on the indoor NYUD2 and outdoor KITTI datasets and analyze the contributions of some key components of our approach. We use the same dataset settings with the second experiment in Sec.~\ref{exp_pretrain}.

\subsubsection{Network comparisons}
In this part, we compare our deep residual network architecture against two baseline networks: deep residual network with 101 and 152 layers in~\cite{kmhe15}. We pretrain the 3 models on our RDIS dataset and finetune on the NYUD2 dataset. Similar to our network architecture, we replace the last 1000-way classification layers of ResNet101 and ResNet152 with one channel convolutional layers during pretraining and 100-way classification layers during finetuning. We also add two convolutional layers with 1024 and 512 channels respectively before the last layer. We do not apply the information gain matrix in this experiment. The results are illustrated in Table~\ref{table:resnet_baseline}. From the table we can see that our network architecture outperforms the deeper ResNet101 and ResNet152.

\begin{table}[h]
	\renewcommand\arraystretch{1.2}
	\caption{Test results on the NYUD2 dataset with different network architectures. The first row is the result of the ResNet101, the second row is the result of the ResNet152, the last row is the result of our network.}
	\centering
	\label{table:resnet_baseline}
	\begin{tabular}{@{\hskip 0.19cm}c@{\hskip 0.11cm}c@{\hskip 0.11cm}c@{\hskip 0.11cm}c @{\hskip 0.39cm}c@{\hskip 0.15cm}c@{\hskip 0.15cm}c}
        \noalign{\smallskip}
		\hline
		\noalign{\smallskip}
						& \multicolumn{3}{c}{\small Accuracy}    & \multicolumn{3}{c}{\small Error}  \\
		                &\small $\delta<1.25$   &\small $\delta<1.25^2$   &\small $\delta<1.25^3$  &\small rel  &\small log10  &\small rms  \\
		\noalign{\smallskip}
		\hline
		\noalign{\smallskip}
    \small Res101   &\small 76.1\%      &\small 94.7\%   &\small 98.5\%   &\small 0.170   &\small 0.071   &\small 0.632  \\
		\noalign{\smallskip}
    \small Res152   &\small 76.2\%      &\small 94.9\%   &\small 98.5\%   &\small 0.169   &\small 0.070   &\small 0.626  \\
		\noalign{\smallskip}
    \small  Ours    &\small \bf 77.8\%   &\small \bf 95.3\%   &\small \bf 98.8\%   &\small \bf 0.159    &\small \bf 0.066    &\small \bf 0.606  \\
		\hline
	\end{tabular}
\end{table}
\subsubsection{Benefit of information gain matrix}

In this part, we evaluate the contribution of the information gain matrix during finetuning. We pretrain the network on our RDIS dataset and finetune on both the NYUD2 and KITTI datasets with and without the information gain matrix. The parameter $\alpha$ defined in Eq.~(\ref{logloss}) is set to $2.0$ and $0.2$ respectively for the NYUD2 and KITTI datasets. The results are illustrated in Table~\ref{table:info_gain}. As we can see from the table that the information gain matrix improves the performance of both indoor and outdoor depth estimation.

\begin{table}
	\renewcommand\arraystretch{1.2}
	\caption{Test results on the NYUD2 and KITTI datasets with and without information gain matrix. For each dataset, the first row is the result without information gain matrix, the following row is the result with information gain matrix.}
	\centering
	\label{table:info_gain}
	\begin{tabular}{@{\hskip 0.05cm}c@{\hskip 0.07cm}c@{\hskip 0.07cm}c@{\hskip 0.07cm}c @{\hskip 0.39cm}c@{\hskip 0.15cm}c@{\hskip 0.15cm}c}
        \noalign{\smallskip}
		\hline
		\noalign{\smallskip}
						& \multicolumn{3}{c}{\small Accuracy}    & \multicolumn{3}{c}{\small Error}  \\
		                 & \small $\delta<1.25$   &\small $\delta<1.25^2$    &\small $\delta<1.25^3$    &\small rel    &\small log10  &\small rms  \\
		\noalign{\smallskip}
		\hline
		\noalign{\smallskip}
		& \multicolumn{5}{c}{\small NYUD2}     \\
		\hline
		\noalign{\smallskip}
   \small Plain    &\small 77.8\%   &\small 95.3\%   &\small 98.8\%   &\small 0.159    &\small 0.066    &\small 0.606  \\
		\noalign{\smallskip}
   \small Infogain    &\small \bf 78.1\%   &\small \bf 95.4\%   &\small \bf 98.9\%   &\small \bf 0.157    &\small \bf 0.065    &\small \bf 0.604  \\
		\hline
		\noalign{\smallskip}
		& \multicolumn{5}{c}{\small KITTI}     \\
		\hline
		\noalign{\smallskip}
   \small Plain    &\small 80.5\%   &\small 93.9\%   &\small 97.7\%   &\small 0.158    &\small 0.064    &\small 5.415  \\
		\noalign{\smallskip}
   \small Infogain   &\small \bf 82.9\%   &\small \bf 94.3\%   &\small \bf 98.2\%   &\small \bf 0.142    &\small \bf 0.058    &\small \bf 5.066  \\
		\hline
	\end{tabular}
\end{table}

\begin{table*}
	\renewcommand\arraystretch{1.2}
	\caption{Comparison with state-of-the-art results on the NYUD2 dataset. The first 7 rows are the results by recent depth estimation methods, the last row is the result by our approach.}
	\centering
	\label{table:stat-of-art_nyu}
	\begin{tabular}{@{\hskip 0.32cm}c@{\hskip 0.25cm}c@{\hskip 0.25cm}c@{\hskip 0.25cm}c @{\hskip 0.6cm}c@{\hskip 0.30cm}c@{\hskip 0.30cm}c}
        \noalign{\smallskip}
		\hline
		\noalign{\smallskip}
						& \multicolumn{3}{c}{\small Accuracy}    & \multicolumn{3}{c}{\small Error}  \\
		                &\small $\delta<1.25$   &\small $\delta<1.25^2$   &\small $\delta<1.25^3$  &\small rel  &\small log10  &\small rms  \\
		\noalign{\smallskip}
		\hline
		\noalign{\smallskip}
    \small Wang et al.~\cite{Wang_2015_CVPR}   &\small 60.5\%    &\small 89.0\%   &\small 97.0\%   &\small 0.210   &\small 0.094   &\small 0.745  \\
		\noalign{\smallskip}
    \small Liu et al.~\cite{LiuSLR15}    &\small 65.0\%    &\small 90.6\%   &\small 97.6\%   &\small 0.213   &\small 0.087   &\small 0.759  \\
		\noalign{\smallskip}
    \small Eigen et al.~\cite{Eigen15}  &\small 76.9\%    &\small 95.0\%   &\small 98.8\%   &\small 0.158   &\small -       &\small 0.641  \\
    		\noalign{\smallskip}
    \small  Laina et al.~\cite{laina2016deeper} &\small 81.1\%    &\small 95.3\%   &\small 98.8\%   &\small  0.127   &\small  0.055  &\small 0.573  \\
    		\noalign{\smallskip}
    \small Li et al.~\cite{junli_depth} &\small 78.9\%    &\small 95.5\%   &\small 98.8\%   &\small  0.152  &\small  0.064  &\small 0.611  \\
			\noalign{\smallskip}
	\small Chakrabarti et al.~\cite{NIPS2016_6510} &\small 80.6\%    &\small 95.8\%   &\small 98.7\%   &\small  0.149  &\small - &\small 0.620  \\
			\noalign{\smallskip}
	\small Xu et al.~\cite{danxu_2017} &\small 81.1\%    &\small 95.4\%   &\small 98.7\%   &\small \bf 0.121  &\small \bf 0.052 &\small 0.586  \\
    		\noalign{\smallskip}
    \small  Ours       &\small \bf 83.1\%    &\small \bf 96.2\%   &\small \bf 98.8\%   &\small 0.132   &\small  0.057  &\small \bf 0.538  \\
		\hline
	\end{tabular}
\end{table*}

\begin{table*}
	\renewcommand\arraystretch{1.2}
	\caption{Comparison with state-of-the-art results on the KITTI dataset. We cap the maximum depth to 50 and 80 meters to compare with recent works. For the work in~\cite{godard2016unsupervised}, we also report their results with additional training images in the CityScapes dataset~\cite{Cordts2016Cityscapes} and denote as Godard et al. CS.}
	\centering
	\label{table:stat-of-art_kitti}
	\begin{tabular}{@{\hskip 0.32cm}c@{\hskip 0.25cm}c@{\hskip 0.25cm}c@{\hskip 0.25cm}c @{\hskip 0.6cm}c@{\hskip 0.30cm}c@{\hskip 0.30cm}c}
        \noalign{\smallskip}
		\hline
		\noalign{\smallskip}
						& \multicolumn{3}{c}{\small Accuracy}    & \multicolumn{3}{c}{\small Error}  \\
		                &\small $\delta<1.25$   &\small $\delta<1.25^2$   &\small $\delta<1.25^3$  &\small rel  &\small rmslog  &\small rms  \\
		\noalign{\smallskip}
		\hline
		\noalign{\smallskip}
		& \multicolumn{5}{c}{\small Cap 80 meters}     \\
		\hline
		\noalign{\smallskip}
    \small Liu et al.~\cite{LiuSLR15}       &\small 65.6\%    &\small 88.1\%   &\small 95.8\%   &\small 0.217   &\small -   &\small 7.046  \\
		\noalign{\smallskip}
    \small Eigen et al.~\cite{EigenPF14}     &\small 69.2\%    &\small 89.9\%   &\small 96.7\%   &\small 0.190   &\small 0.270   &\small 7.156  \\
		\noalign{\smallskip}
    \small Godard et al.~\cite{godard2016unsupervised}    &\small 81.8\%    &\small 92.9\%   &\small 96.6\%   &\small 0.141   &\small 0.242   &\small 5.849  \\
		\noalign{\smallskip}
    \small Godard et al. CS~\cite{godard2016unsupervised}   &\small 83.6\%    &\small 93.5\%   &\small 96.8\%   &\small 0.136   &\small  0.236  &\small 5.763  \\
    		\noalign{\smallskip}
    \small  Ours     &\small \bf 89.0\%    &\small \bf 96.7\%   &\small \bf 98.4\%   &\small \bf 0.120   &\small \bf 0.192  &\small \bf 4.533  \\
		\hline
		\noalign{\smallskip}
		& \multicolumn{5}{c}{\small Cap 50 meters}     \\
		\hline
		\noalign{\smallskip}
    \small  Garg  et al.~\cite{garg2016unsupervised}   &\small 74.0\%    &\small 90.4\%   &\small 96.2\%   &\small 0.169   &\small  0.273  &\small 5.104  \\
		\noalign{\smallskip}
    \small  Godard et al.~\cite{godard2016unsupervised}   &\small 84.3\%    &\small 94.2\%   &\small 97.2\%   &\small 0.123   &\small  0.221  &\small 5.061  \\
    		\noalign{\smallskip}
    \small  Godard et al. CS~\cite{godard2016unsupervised}  &\small 85.8\%    &\small 94.7\%   &\small 97.4\%   &\small 0.118   &\small  0.215  &\small 4.941  \\
    		\noalign{\smallskip}
    \small  Ours  &\small \bf 89.7\%   &\small \bf 96.8\%   &\small \bf 98.4\%   &\small \bf 0.117   &\small \bf 0.189  &\small \bf 3.753  \\
		\hline
	\end{tabular}
\end{table*}

\begin{table}[h]
	\renewcommand\arraystretch{1.2}
	\caption{Test results of depth estimation by classification and regression on the NYUD2 and KITTI datasets. For each dataset, the first row is the result of regression, the following row is the result of classification.}
	\centering
	\label{table:classi_vs_regress}
	\begin{tabular}{@{\hskip 0.19cm}c@{\hskip 0.11cm}c@{\hskip 0.11cm}c@{\hskip 0.11cm}c @{\hskip 0.39cm}c@{\hskip 0.15cm}c@{\hskip 0.15cm}c}
        \noalign{\smallskip}
		\hline
		\noalign{\smallskip}
						& \multicolumn{3}{c}{\small Accuracy}    & \multicolumn{3}{c}{\small Error}  \\
		                &\small $\delta<1.25$   &\small $\delta<1.25^2$   &\small $\delta<1.25^3$  &\small rel  &\small log10  &\small rms  \\
		\noalign{\smallskip}
		\hline
		\noalign{\smallskip}
		& \multicolumn{5}{c}{\small NYUD2}     \\
		\hline
		\noalign{\smallskip}
    \small regression   &\small 66.9\%      &\small 92.1\%   &\small 98.0\%   &\small 0.215   &\small 0.084   &\small 0.730 \\
		\noalign{\smallskip}
    \small classification  &\small \bf 72.3\%   &\small \bf 92.7\%   &\small \bf 98.3\%   &\small \bf 0.195    &\small \bf 0.077    &\small \bf 0.691  \\
		\hline
		\noalign{\smallskip}
		& \multicolumn{5}{c}{\small KITTI}     \\
		\hline
		\noalign{\smallskip}
		\small regression   &\small 68.9\%      &\small 89.4\%   &\small 91.1\%   &\small 0.256   &\small 0.092   &\small 7.160 \\
		\noalign{\smallskip}
    \small classification  &\small \bf 79.9\%   &\small \bf 93.7\%   &\small \bf 97.6\%   &\small \bf 0.166    &\small \bf 0.067    &\small \bf 5.443  \\
		\hline
	\end{tabular}
\end{table}

\begin{table}
	\renewcommand\arraystretch{1.35}
	\newcommand{\tabincell}[2]{\begin{tabular}{@{}#1@{}}#2\end{tabular}}
	\caption{Comparison with state-of-the-art results on the DIW dataset. The evaluation metric is Weighted Human Disagreement Rate (WHDR).}
	\centering
	\label{table:ordinal_compare}
	\begin{tabular}{@{\hskip 0.25cm}c@{\hskip 0.35cm}|c@{\hskip 0.35cm}c@{\hskip 0.39cm}c}
        \noalign{\medskip}
		\hline
     \small Method   &\small  WHDR   \\
  		\hline
    \small Baseline~\cite{DIW_NIPS2016}        &\small 31.37\%    \\
    \small Eigen~\cite{DIW_NIPS2016}           &\small 25.70\%    \\
    \small Chen-NYU~\cite{DIW_NIPS2016}      &\small 31.31\%    \\
    \small Chen-DIW~\cite{DIW_NIPS2016}      &\small 22.14\%    \\
    \small Chen-NYU-DIW~\cite{DIW_NIPS2016}      &\small 14.39\%    \\
    \small Ours-RDIS                            &\small 18.05\%    \\
    \small Ours-NYU-RDIS                           &\small \bf 11.55\%    \\
		\hline
	\end{tabular}
\end{table}

\subsubsection{Depth classification vs. depth regression}
In this part, we compare our depth estimation by classification with the conventional regression. We directly train the ResNet101 model without pretraining on our RDIS dataset. For depth regression, we use the $L2$ loss. For our depth estimation as classification, we discretize the continuous depth values into 100 and 50 bins in the log space respectively for the NYUD2 and KITTI datasets. And we set the parameter $\alpha$ to $2.0$ and $0.2$ respectively. We show the results in Table~\ref{table:classi_vs_regress}, from which we can see that our depth estimation by classification outperforms the conventional regression. This is because the regression tends to converge to the mean depth values, which may cause larger errors in areas that are either very far from or very close to the camera. The classification naturally produces the confidence of a depth estimation in the form of probability distribution. Based on the probability distribution, we can apply the information gain loss to alleviate the problem.

\begin{figure*}
	\begin{center}
		\includegraphics[scale=.52]{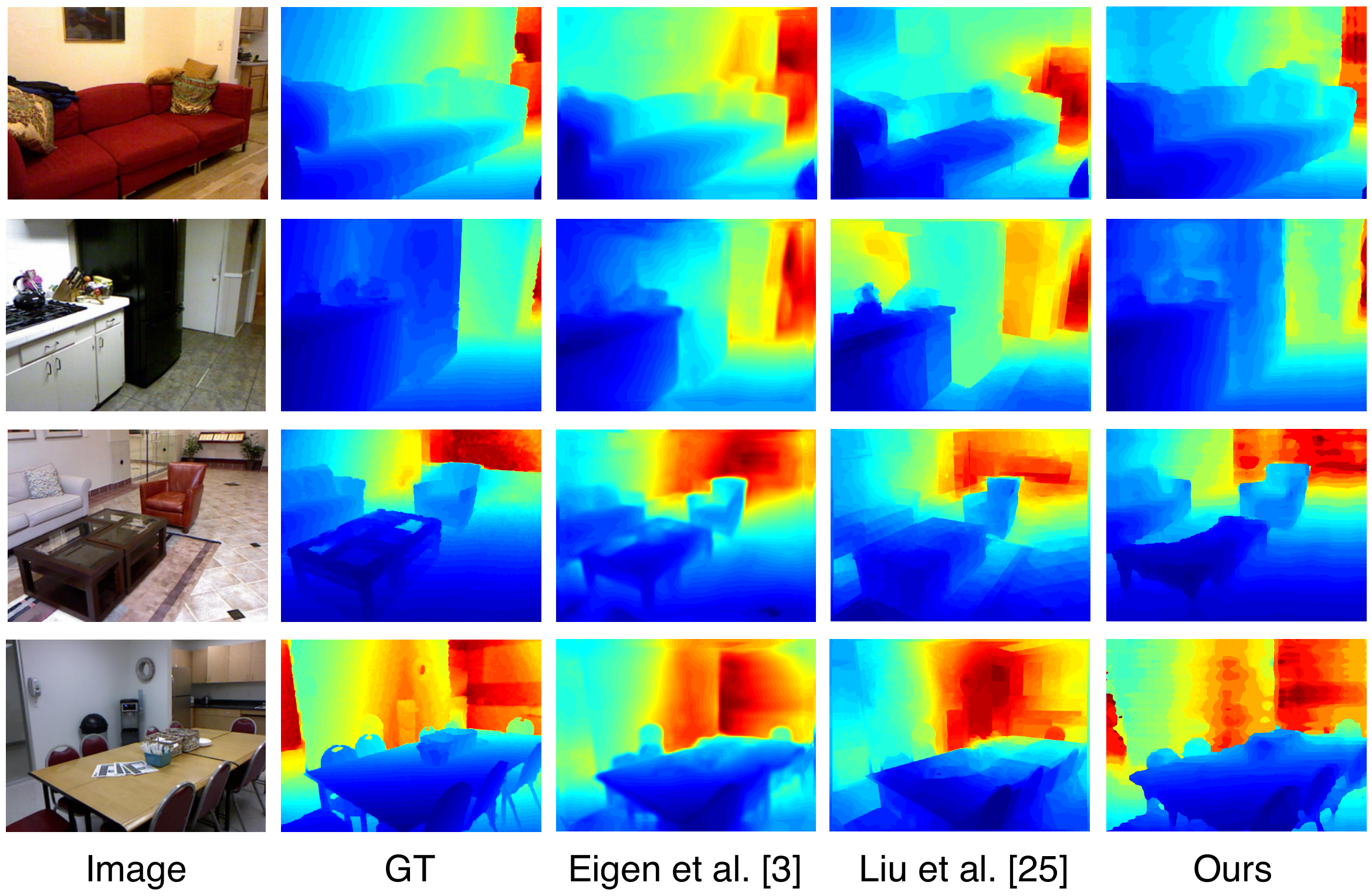}
	\end{center}
	\caption{Qualitative comparisons with state-of-the-art results on the NYUD2 dataset. The first two columns are RGB images and ground-truth depths respectively. The following 4 columns are predictions. Depths are shown in color (red is far, blue is close).}
	\label{fig_nyusample}
\end{figure*}

\begin{figure*}
	\begin{center}
		\includegraphics[scale=.45]{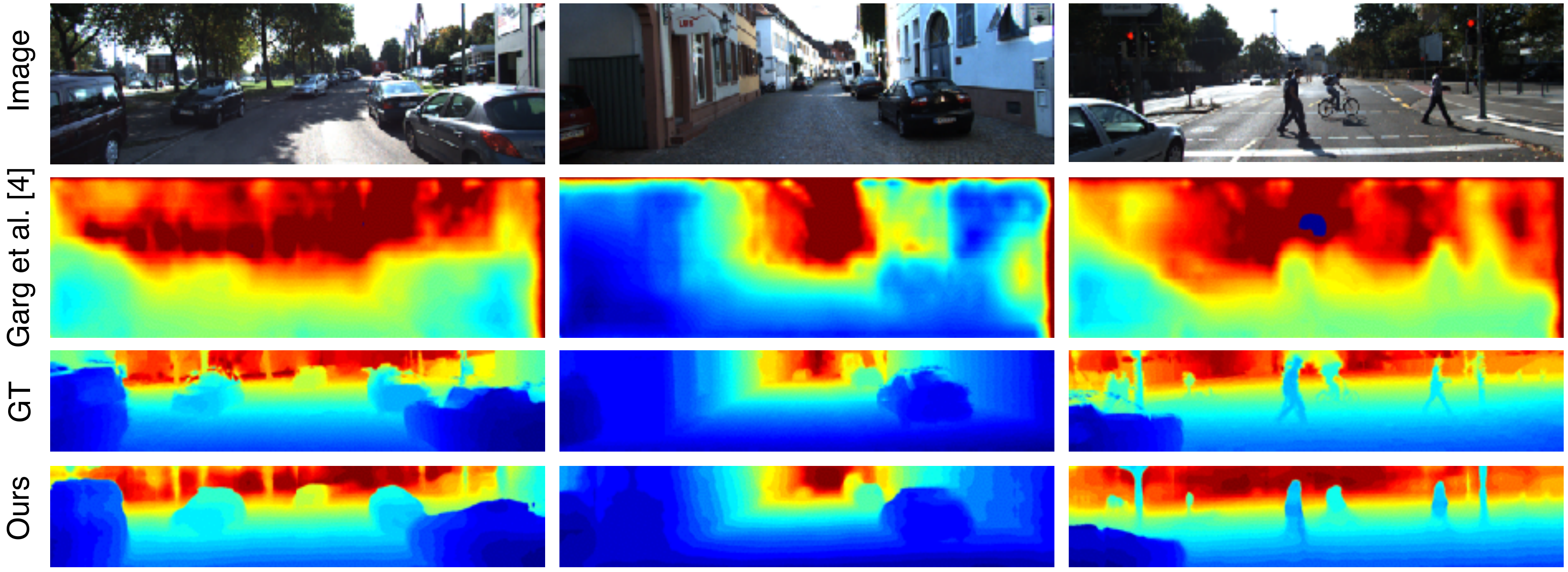}
	\end{center}
	\caption{Qualitative comparisons with state-of-the-art results on the KITTI dataset. Depths are shown in color (red is far, blue is close). Since the ground-truth captured by the velodyne is very sparse, we inpaint the ground-truth for visualization purposes. We also crop the ground-truth and our predictions to mask out the vast sky regions.}
	\label{kitti_sample}
\end{figure*}

\begin{figure*}
	\begin{center}
		\includegraphics[scale=.52]{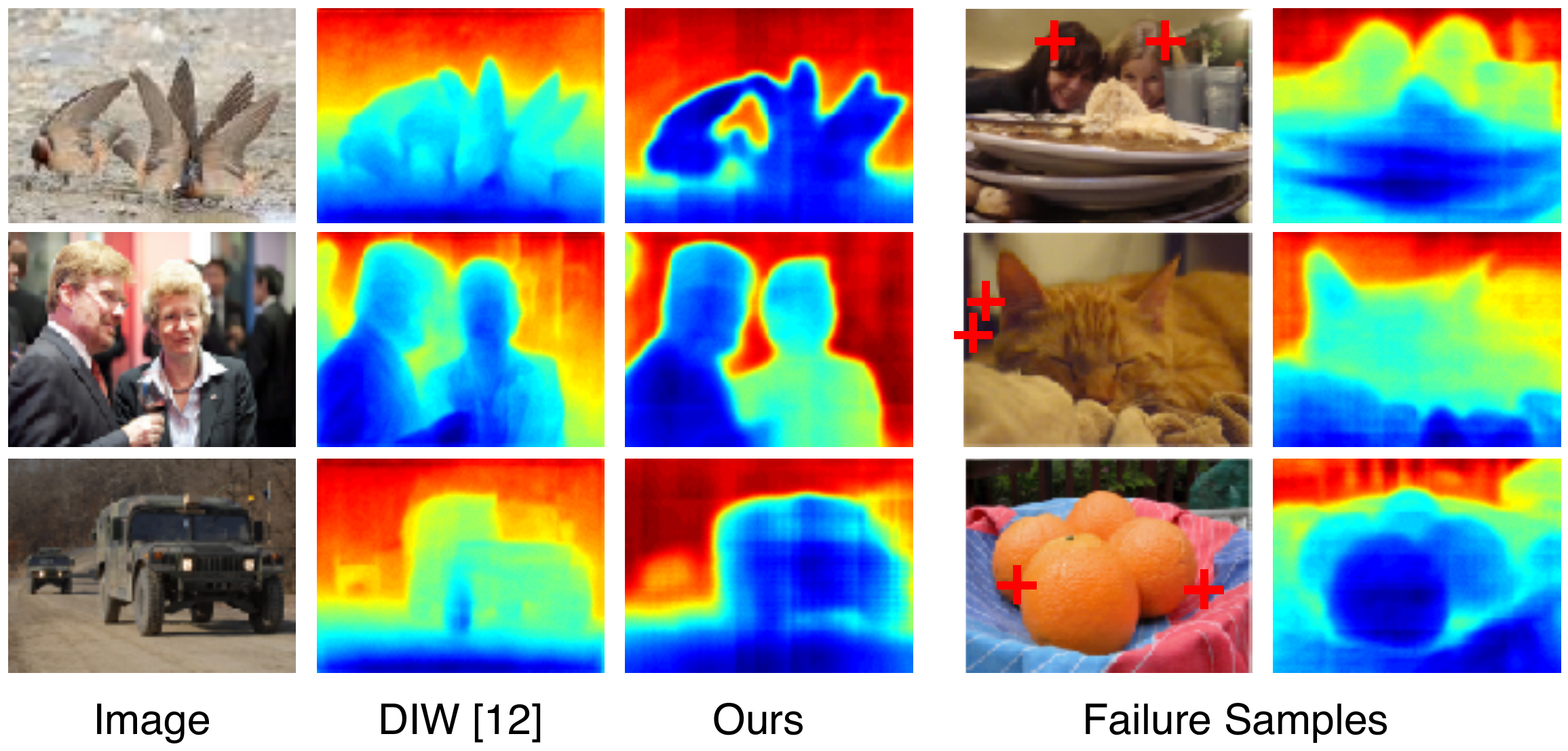}
	\end{center}
	\caption{Some examples of relative depth estimation of the DIW dataset. The first column are the RGB images, the second column are the predictions of Chen et al.~\cite{DIW_NIPS2016}, the third column are our predictions. The last two columns are some failure samples of our approach. The pairs of points labelled with ground-truth ordinal relations are marked as red crosses. }
	\label{fig_DIWsample}
\end{figure*}

\subsection{State-of-the-art comparisons}

In this section, we evaluate metric depth estimation on the NYUD2 and KITTI datasets and compare with recent depth estimation methods. During pretraining, we use 1K pairs of points in each input image. During finetuning, we discretize the continuous metric depths into 100 and 50 bins in log space for the NYUD2 and KITTI datasets respectively. We also evaluate relative depth estimation on the Depth in the Wild (DIW)~\cite{DIW_NIPS2016} dataset.

\subsubsection{NYUD2}
We finetune our model on the raw NYUD2 training set and test on the standard 654 images. We set the parameter $\alpha$ of the information gain matrix to be $2.0$. We compare our approach against several prior works and report the results in Table \ref{table:stat-of-art_nyu}, from which we can see that we achieve state-of-the-art results of 4 evaluation metrics without using any multi-scale network architecture, up-sampling or CRF post-processing. Fig.~\ref{fig_nyusample} illustrates some qualitative evaluations of our method compared against Liu et al.~\cite{LiuSLR15} and Eigen et al.~\cite{Eigen15}.

\subsubsection{KITTI}

We finetune our model on the same training set in \cite{godard2016unsupervised} which contains 33131 images and test on the same 697 images in \cite{EigenPF14}. But different with the depth estimation method proposed in \cite{godard2016unsupervised} which applies both the left and right images in stereo pairs, we only use the left images. The missing values in the ground-truth depth maps are ignored during finetuning and evaluation. We set the parameter $\alpha$ of the information gain matrix to be $0.2$. In order to compare with the recent state-of-the-art results, we cap the maximum depth to both 80 meters and 50 meters and present the results in Table~\ref{table:stat-of-art_kitti}. We outperform state-of-the-art results of all evaluation metrics significantly. Some qualitative results are illustrated in Fig.~\ref{kitti_sample}. Our method yields outstanding visual predictions.

\subsubsection{DIW}
We evaluate relative depth estimation on the DIW test set and report the WHDR of 7 methods in Table~\ref{table:ordinal_compare}: 1) a baseline method that uses only the location of the query points: classify the lower point to be closer or guess randomly if the two points are at the same height (Baseline); 2) the model trained by Eigen et al.~\cite{Eigen15} on the raw NYUD2 dataset (Eigen); 3) the model trained by Chen et al.~\cite{DIW_NIPS2016} on the raw NYUD2 dataset (Chen-NYU); 4) the model trained by Chen et al.~\cite{DIW_NIPS2016} on the DIW dataset (Chen-DIW). 5) the model by Chen et al.~\cite{DIW_NIPS2016} pretrained on the raw NYUD2 dataset and finetuned on the DIW dataset (Chen-NYU-DIW). 6) our model trained on our RDIS dataset (Ours-RDIS). 7) our model pretrained on the raw NYUD2 dataset and finetuned on our RDIS dataset (Ours-NYU-RDIS). From the table we can see that even though we do not train our model on the DIW training set, we achieve state-of-the-art result on the DIW test set. We show some of our predicted relative depth maps as well as some failure samples in Fig.~\ref{fig_DIWsample}, from which we can see that our predicted relative depth maps are visually better. As for the failure samples, we can also predict satisfactory relative depth maps. Notably, the ground-truth pairs in failure samples are those points with almost equal distance. Given the fact that the equal relation is absent in DIW, we can conclude that we reach the nearly perfect performance on the DIW test set.

\section{Conclusion}
We have proposed a new dataset Relative Depths in Stereo (RDIS) containing images labelled with dense relative depths. The ground-truth relative depths are obtained through existing stereo algorithm as well as manual post-processing. We have shown that augmenting benchmark RGB-D datasets with our proposed RDIS dataset, the performance of single-image depth estimation can be improved significantly.

Note that the goal of this work is to predict depths from single monocular images. However the application of our proposed RDIS dataset is not limited to this. With the learning scheme based on relative depths, we can perform 2D-to-3D conversion like Deep3D \cite{xie2016deep3d}. We leave this as our future work.

\ifCLASSOPTIONcaptionsoff
\newpage
\fi

\bibliographystyle{IEEEtran}
\end{document}